%% file: 0-main.tex
\documentclass[conference]{IEEEtran}
\usepackage[usenames,dvipsnames,table,xcdraw]{xcolor}

\newcommand{\algname}{Sliding and Grasping for Tangle Manipulation (SGTM)}

\newcommand{\algabbr}{SGTM}

\newcommand{\todo}[1]{}
\renewcommand{\todo}[1]{{\color{red} TODO: {#1}}}
\usepackage[numbers]{natbib}
\usepackage{mathtools}
\let\labelindent\relax
\usepackage[shortlabels]{enumitem}
\usepackage{caption}
\usepackage{amsmath, amssymb, amscd}
\usepackage{algorithm}
\usepackage[noend]{algpseudocode}
\usepackage[bookmarks=true]{hyperref}

\begin{document}

\title{Autonomously Untangling Long Cables}


\author{Vainavi Viswanath$^{*1}$, Kaushik Shivakumar$^{*1}$, Justin Kerr$^{*1}$, Brijen Thananjeyan$^{1}$, 
Ellen Novoseller$^{1}$, \\ 
Jeffrey Ichnowski$^{1}$, Alejandro Escontrela$^{1, 3}$, Michael Laskey$^{2}$, Joseph E. Gonzalez$^{1}$, Ken Goldberg$^{1}$}

\makeatletter
\def\thanks#1{\protected@xdef\@thanks{\@thanks \protect\footnotetext{#1}}}
\makeatother

\thanks{$^{1}$AUTOLAB at the University of California, Berkeley}
\thanks{$^{2}$Toyota Research Institute}
\thanks{$^{3}$Google Brain}
\thanks{$^{*}$equal contribution}



%

\maketitle

\noindent
\textit{Note: This is a slightly revised version of the Best Systems Paper Award paper from RSS 2022.}\\

\begin{abstract}

Cables are ubiquitous in many settings and it is often useful to untangle them. However, cables are prone to self-occlusions and knots, making them difficult to perceive and manipulate. The challenge increases with cable length: long cables require more complex slack management to facilitate observability and reachability. In this paper, we focus on autonomously untangling cables up to 3 meters in length using a bilateral robot. We develop RGBD perception and motion primitives to efficiently untangle long cables and novel gripper jaws specialized for this task. We present \algname{}, an algorithm that composes these primitives to iteratively untangle cables with success rates of 67\% on isolated overhand and figure-eight knots and 50\% on more complex configurations. Supplementary material, visualizations, and videos can be found at \texttt{\url{https://sites.google.com/view/rss-2022-untangling/home}}.

\end{abstract}

\IEEEpeerreviewmaketitle

\input{1-introduction.tex}
\input{2-related-work.tex}
\input{3-problem-statement.tex}

\input{4-methods.tex}
\input{5-experiments.tex}
\input{6-conclusion.tex}
\input{7-acknowledgements}

\bibliographystyle{plainnat}
\bibliography{references}

\end{document}

%% file: 1-introduction.tex
\section{Introduction}
\label{sec:intro}

Long cables are commonplace in household and industrial settings, from  wires in homes to cables in manufacturing plants to ropes in sailing~\cite{mayer2008system, sanchez2018robotic,  van2010superhuman, yamakawa2007one}. Cables---defined here as single-dimensional deformable objects---often tangle and form knots, which can be unsightly, unsafe, and reduce utility. However, autonomously manipulating cables to untangle them is challenging due to their infinite-dimensional configuration spaces and tendency to form self-occlusions and knots~\cite{grannen2020untangling,sundaresan2021untangling,viswanath2021disentangling,sundaresan2020learning,lui2013tangled}. 
These challenges grow with increasing cable length as longer cables allow more knots to form. Additionally, as robots untangle one segment of a cable, the other cable segments, the \emph{slack}, can form new knots, occlude visibility, or impede grasping.
The stiffness of cables further adds to the slack management challenge, because untangled slack may appear to hold a knot shape even without a knot present. Robustly manipulating long cables requires policies that perceive cable configurations and effectively manage slack during manipulation.

\begin{figure}[ht!]
\centering
\includegraphics[width=1.0\linewidth]{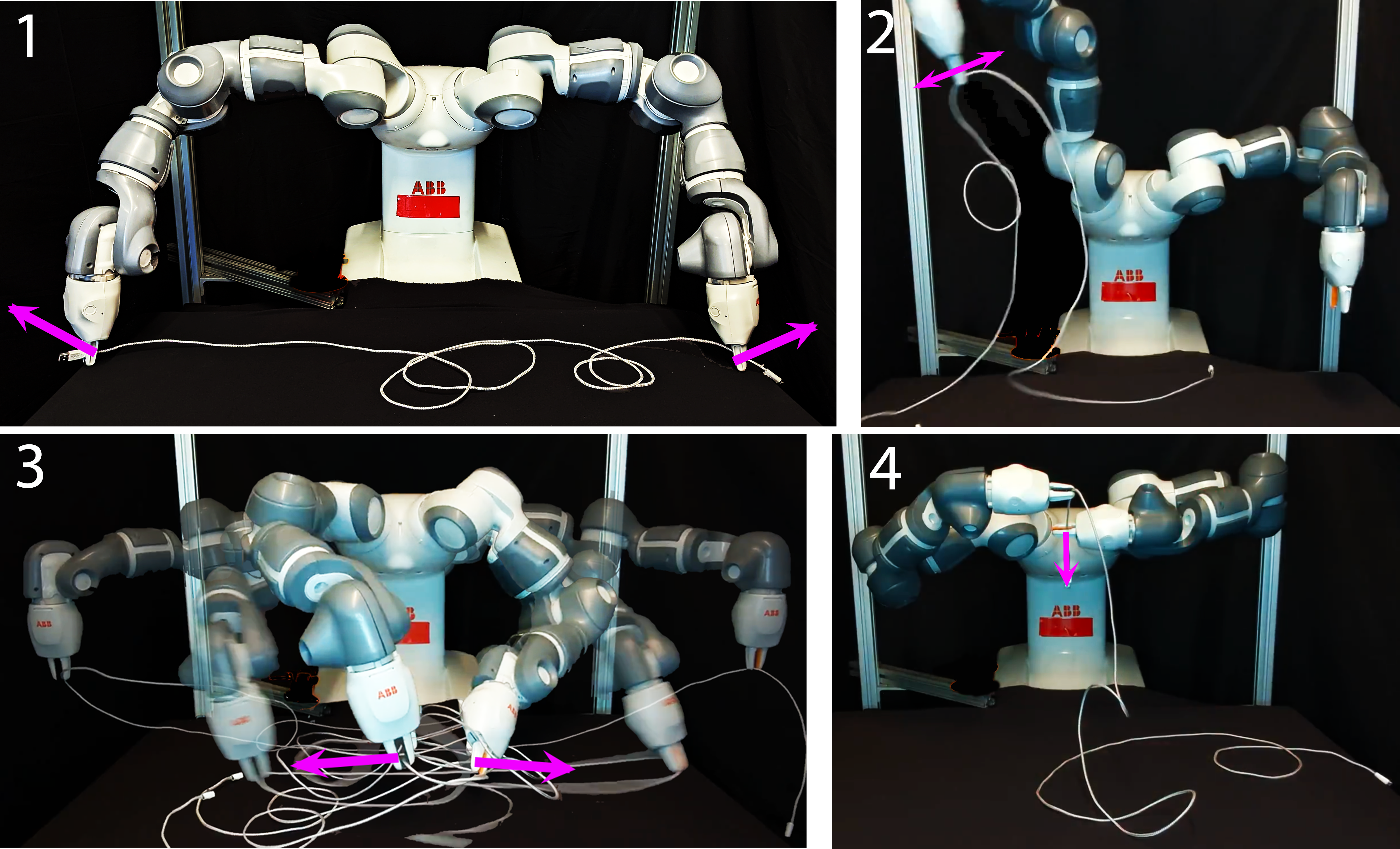}
\caption{\textbf{Overview of \algname{}}:  \algabbr{} untangles a long cable by (1) performing endpoint separation to spread out the configuration, 2) shaking to reduce loops and reveal knots, 3) performing a dual-cage separation primitive to untangle a knot, and (4) performing physical tracing to verify the cable is untangled.}\vspace{-0.25in}
\label{fig:figure1}
\end{figure}

Prior work in robot cable untangling considers short cables and studies how to untangle dense knots, which have no open space between adjacent cable segments~\cite{grannen2020untangling,sundaresan2021untangling,viswanath2021disentangling}, or considers longer cables, estimated to be no longer than one meter in length, whose paths can be clearly traced using analytic methods~\cite{lui2013tangled}. This paper studies untangling knots in cables up to 3\,m long that have sufficient stiffness to prevent the formation of dense knots. Untangling such cables requires policies to actively manage the slack created during manipulation to prevent the formation of new knots and crossings. We use overhead RGBD images for perception and a bilateral YuMi robot.

\begin{figure}[t!]
\centering
\includegraphics[width=1.0\linewidth]{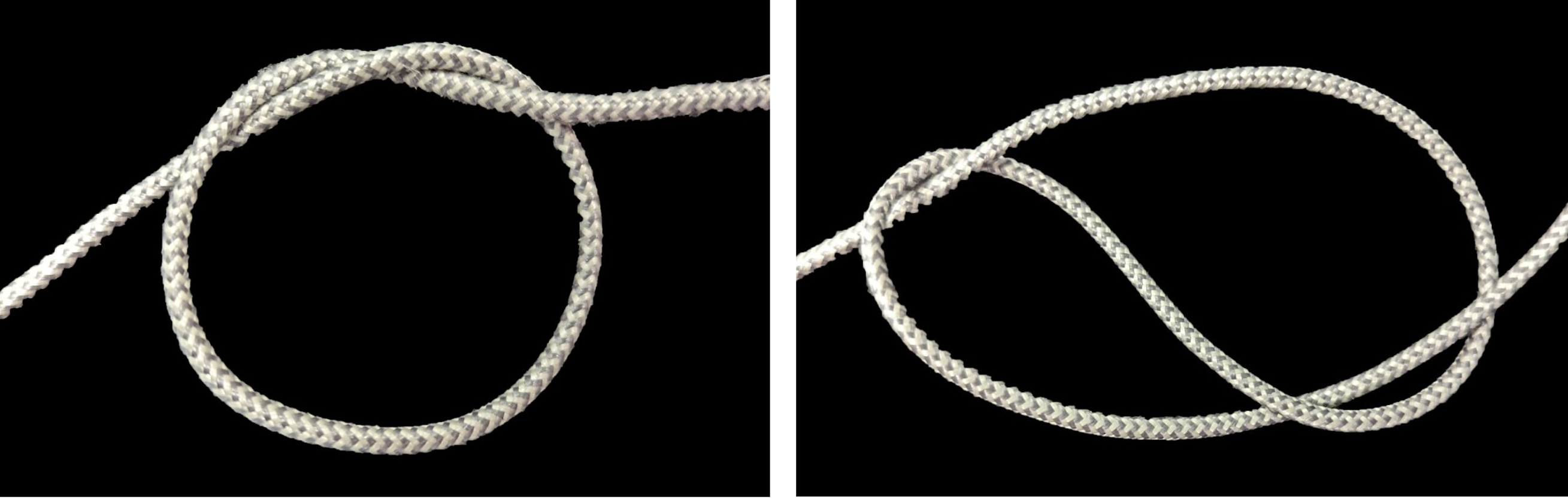}
\caption{\textbf{Knots that \algabbr{} Untangles}: Left: an overhand knot. Right: a figure 8 knot.}
\label{fig:typesofknots}
\end{figure}

\setlength{\belowcaptionskip}{-10pt}
\looseness=-1 This paper makes 3 contributions:
\textbf{(1)} Novel gripper jaws for untangling that enables two modes of grasping: caging and pinching. Caging grasps enable the cable to slide along the gripper jaws without falling out, which is useful for physically tracing along the cable. Pinching grasps firmly hold the cable, preventing it from moving.
\textbf{(2)} Three novel bilateral manipulation primitives for cable untangling: shaking, tracing, and dual-cage separation. Shaking moves can simplify the cable state before attempting other primitives. Physical tracing can slide along the cable to actively uncover knots and loosen dense structures. dual-cage separation actions can loosen and undo knots using learning-based perception to identify isolated knots in the scene. The dual-cage separation action slides apart from a knot center, avoiding the need to pinch and regrasp the cable repeatedly to undo knots as in prior work~\cite{sundaresan2021untangling,grannen2020untangling, viswanath2021disentangling}.
Building on our previous work, this paper contributes a learning-based perception pipeline which leverages RGBD sensing to identify cable endpoints and knots~\cite{viswanath2021disentangling,sundaresan2021untangling,grannen2020untangling}, which are used to select and parameterize the novel manipulation primitives.
\textbf{(3)} A novel untangling algorithm, \algname{} (Fig. ~\ref{fig:figure1}), to untangle cables of up to 3 meters in length from starting configurations that contain up to 2 overhand and figure 8 knots (Fig.~\ref{fig:typesofknots}), as well as self-crossings that do not form knots. To the best of our knowledge, this is also the first work on robot cable untangling that considers cables up to 3 meters with multiple knots.

%% file: 2-related-work.tex
\section{Related Work}

\subsection{Deformable Object Manipulation}

Autonomously manipulating deformable objects such as cables (1D), fabric (2D), and bags (3D) is a challenging problem, as such materials have near-infinite configuration spaces, form self-occlusions, and are subject to complex dynamics. This work focuses on cables, which can present particularly complex configurations due to their tendency to form knots. While recent years have seen significant progress in robot deformable manipulation, for instance perception-driven algorithms for untangling cables~\cite{grannen2020untangling, sundaresan2021untangling, viswanath2021disentangling,lui2013tangled}, smoothing and folding fabric~\cite{seita2019deep, weng2022fabricflownet,ganapathi2020learning,hoque2020visuospatial,kollar2022simnet}, and placing objects into bags~\cite{seita2020learning}, to the best of our knowledge, the task of autonomously untangling long household cables remains unstudied.

Developing effective manipulation algorithms for deformable objects often involves choosing an abstract object representation on which an algorithm can operate. Existing work on deformable manipulation has leveraged a number of approaches for representing deformable objects. For instance, dense descriptors~\cite{florence2018dense}, which map object images to a pixel-wise embedding, have been successfully applied to both cable knot tying~\cite{sundaresan2020learning} and fabric folding and smoothing~\cite{ganapathi2020learning}. Yet, this approach is not robust to the severe deformation and occlusion common with lengthier cables. Other works attempt to address this challenge by learning dynamics models of deformable objects, which can then be used for planning
\cite{hoque2020visuospatial, wang2019learning2, yan2020learning, lin2022learning}.
These methods have positioned non-knotted cables~\cite{yan2020learning, wang2019learning2} and fabric~\cite{hoque2020visuospatial, yan2020learning, lin2022learning} into target configurations. Yet, for heavily self-occluded knotted cables, robust state estimation remains challenging, while dynamics are governed by complex knot structure. 
Instead of estimating the full state of the cable, our work leverages perception-based keypoint prediction methods and knowledge about the geometric problem structure.

Several works meanwhile propose end-to-end model-free visuomotor learning approaches for deformable manipulation. These include model-free reinforcement learning, which has been applied successfully to fabric smoothing and folding~\cite{matas2018sim, wu2019learning, lee2020learning} and straightening unknotted rope~\cite{wu2019learning}, as well as leveraging optical flow to predict fabric manipulation actions given a target image~\cite{weng2022fabricflownet}. Imitation learning approaches, which require demonstrations, have also seen success in manipulating fabric and non-knotted ropes~\cite{seita2019deep, seita2020learning, nair2017combining}. However, these model-free approaches are highly sample inefficient for cable untangling, as they do not leverage geometric knowledge specific to cable structure or untangling.

\subsection{Cable Manipulation and Untangling}
\looseness=-1 Existing works on robot cable untangling do not consider long cables, such as those commonly found in households, which pose additional challenges compared to short cables.
This paper extends several ideas from prior methods, which use a combination of vision-based perception and domain structure to extract the state features necessary for untangling. Firstly,~\citet{lui2013tangled} start from initial configurations in which the cable's self-crossings are distinctly visible and use RGBD sensing to extract the cable's full state via classical feature extraction. The features are used to construct a graphical representation of the cables' self-crossings, which are used for untangling. The cables in this work are of unspecified length but do not appear to be longer than one meter in length. Meanwhile, in~\cite{grannen2020untangling,viswanath2021disentangling,sundaresan2021untangling}, the robot does not perform full state estimation and instead predicts keypoints to parameterize open-loop untangling actions such as crossing removal and pulling the endpoints apart. The robot untangles very densely knotted single and multi-cable knots in short cables without significant slack. Similar to these prior works, \algabbr{} uses RGBD sensing and predicts parameterized actions that do not explicitly require full state estimation. However, while prior works focus on short cables, \algabbr{} is designed for long cables, in which the cable's underlying knot structure can be significantly obfuscated due to the additional slack, which can create a number of self-crossings and self-occlusions that are not part of a knot. 

This work defines novel motion primitives for cable untangling and employs perception-based methods to detect keypoints for executing these primitives. Several of these novel manipulation primitives slide along the length of the cable, which enable the robot to undo knots faster and physically trace the cable to check for knots. Sliding motions in cable manipulation were studied in~\cite{she2019cable} using a tactile gripper with learning-based control to keep the cable centered in the grippers when sliding along its length. Instead of using tactile sensing to fix the cable in the grippers, we propose a physical gripper augmentation that enables two grasping modes without sophisticated control techniques: caging, which enables sliding, and pinching, which constrains all movement in the gripper (Figure~\ref{fig:grippers}). This approach allows for a lower-profile gripper which can fit in tighter knot configurations and can speed up manipulation time due to the absence of active control. In addition, caging grasps enable gripper motion along the cable without applying force to the cable, a mode in which a tactile sensor could not collect feedback.

%% file: 3-problem-statement.tex
\section{Problem Statement}
\label{sec:ps}
\subsection{Workspace Definition and Assumptions}
\label{subsec:workspace_definition}
\looseness=-1 We define an $(x, y, z)$ Cartesian coordinate frame containing a bilateral robot and a flat manipulation surface that is parallel to the $xy$-plane. We assume an overhead RGB-D camera facing the manipulation surface from above and the rigid transformations between the camera, robot, and workspace coordinate frames are known. Because the RGBD camera uses structured light, we assume the workspace must be static during image capture.

At each iteration of a rollout ($t \in 0, 1, ...N$), the manipulation workspace contains an incompressible cable $\mathcal{C}$ with cross-section radius $r$ and length $l$, which traces a continuous volume $c_t(s): [0, 1] \rightarrow (x, y, z)$ in the workspace. The parameter $s \in [0, 1]$ is used to index into the cable path (with length normalized to 1), with $c_t(s)$ corresponding to the center of the circular cable cross-section with distance $sl$ from the endpoint defined as the first endpoint. The circular slice lies in the plane orthogonal to the cable's local direction, $\nabla_s c_t(s)$.

We assume that the cable can be segmented from the manipulation surface using color thresholding and that in the starting configuration $c_0$, the entire cable is contained in the workspace. The initial cable configuration can form self-crossings, which could include knots or loops. A cable segment contains at least one \textit{knot} if pulling it taut by pulling the two endpoints in opposite directions does not result in a straight cable, i.e. one with no crossings. We assume that any knots in the initial configuration are either overhand or figure-eight knots (Fig.~\ref{fig:typesofknots}), and that sufficient space within each knot exists to fit both robot jaws inside. We assume that individual knots are distinct, such that no knots are embedded within other knots. We classify the initial cable configurations we consider into tiers of difficulty in Section~\ref{sec: experiments}.

\begin{figure}[t!]
\centering
\includegraphics[width=0.8\linewidth]{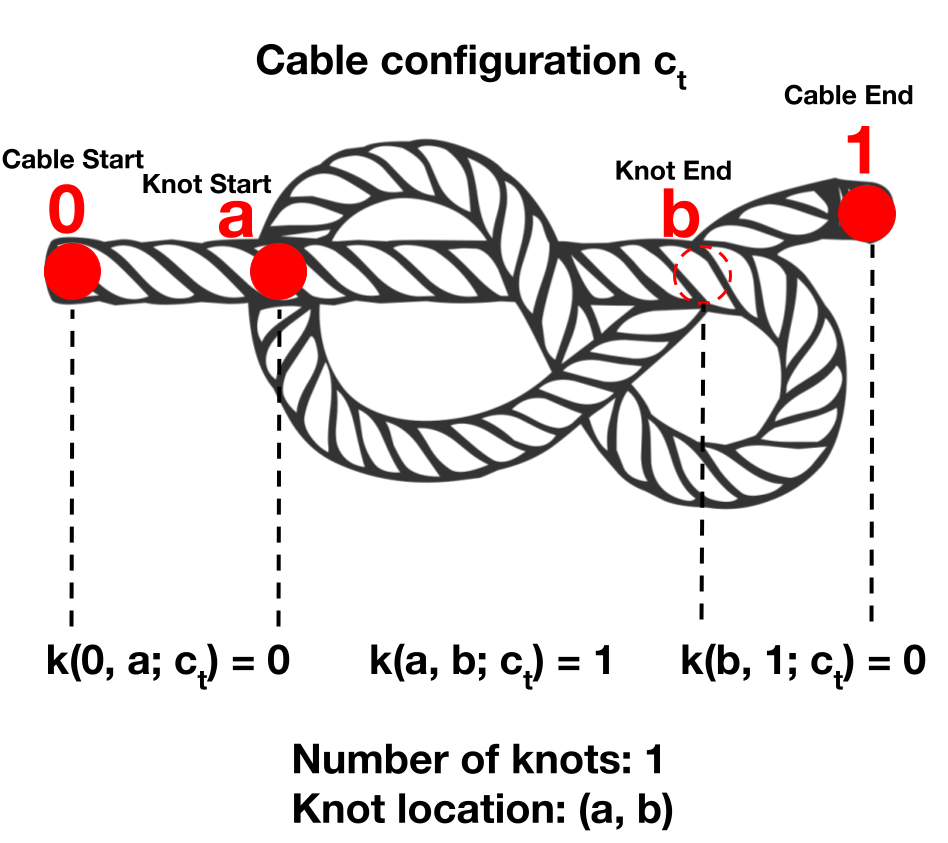}
\caption{\textbf{Knot Definition on Figure-Eight Knot}: for a cable configuration at time $t$, $c_t$, a knot exists between two points $(a, b)$ if pulling apart at those points will not result in a straight cable. In this example, no knots exist in $(0, a)$ and $(b, 1)$; thus, $k(0, a; c_t) = 0$ and $k(b, 1; c_t) = 0$. A figure-eight knot exists between indices $a$ and $b$, and so $k(a, b; c_t) = 1$. Observe that $(a, b)$ is the smallest interval that contains the knot, such that $\mathcal{K}(c_t) = \left\{(a, b)\right\}$.
}
\label{fig:knot_def}
\end{figure}

\subsection{Knot Definition}\label{subsec:knot_definition}
Let $a, b \in [0, 1]$ be two indices on the cable's path, where $a < b$. We say that a knot exists at time $t$ between indices $a$ and $b$ if grasping and pulling apart the points $c_t(a)$ and $c_t(b)$ does not straighten the cable segment (i.e., by removing all crossings) between $a$ and $b$. We let $k(a,b; c_t) = 1$ if a knot exists between $a$ and $b$ on cable $c_t$, otherwise $k(a,b; c_t) = 0$. For a particular cable configuration $c_t$, we can represent its knot structure by the minimal set of intervals $\mathcal{K}(c_t) = \{(a_1,b_1),\ldots,(a_n,b_n)\}$ containing a knot:

\begin{align*}
    \mathcal{K}(c_t)&= \text{argmin}_{\mathcal{K}(c_t) \in \bigcup_{k=0}^\infty([0, 1]\times[0, 1])^k} \sum_{(a_i, b_i)\in \mathcal{K}(c_t)} |b_i - a_i|\\
    \text{s.t.}\quad &k(a_i, b_i; c_t) = 1,\ \forall i \in \{1, \ldots, |\mathcal{K}(c_t)|\}, \\ & a_i < b_i\ \forall \ i; \,\,\, b_i \le a_{i + 1} \ \forall \ i, \\ & k(\tilde{a}, \tilde{b}; c_t) = 0 \
    \forall \  \tilde{a}, \tilde{b}
    \\ & \text{ s.t. } \ b_i \le \tilde{a} \le \tilde{b} \le a_{i + 1}, \ b_0 := 0, a_{|\mathcal{K}(c_t)|+1} := 1.
\end{align*}

This definition finds the smallest interval surrounding each distinct knot in the cable. Specifically, it enforces that a knot exists between each $(a_i, b_i)$ pair and also that no knots are excluded, meaning $k(\tilde{a}, \tilde{b}; c_t) = 0$ between successive knots. For a given cable configuration $c_t$, the number of distinct knots is the number of intervals in $\mathcal{K}(c_t)$. This notation is illustrated in Figure~\ref{fig:knot_def} for a single figure-eight knot on a cable.

\subsection{Task Objective}
The goal of the cable untangling problem is to manipulate the cable into a configuration $c_t$ such that $\mathcal{K}(c_t) = \emptyset$, indicating that no knots exist in the cable. We call such a configuration \textit{untangled}.

\subsubsection{Algorithm Inputs and Outputs}
The algorithm is provided with overhead RGBD images as input at each iteration and it outputs either an open-loop trajectory to execute with the bilateral robot or a termination signal. To output a termination signal, the robot must detect that it has successfully untangled the cable.


\subsubsection{Performance Metrics}
We record the success rate across three difficulty tiers of problem instances, as described in Section~\ref{sec: experiments}. We also report the time taken for the algorithm to terminate in each trial, as well as the time taken to first reach an untangled state, which may differ from the total time if the algorithm takes additional time to identify that it has fully untangled the cable.


%% file: 4-methods.tex
\section{Methods}

\begin{figure}[t!]
\centering
\includegraphics[width=0.95\linewidth]{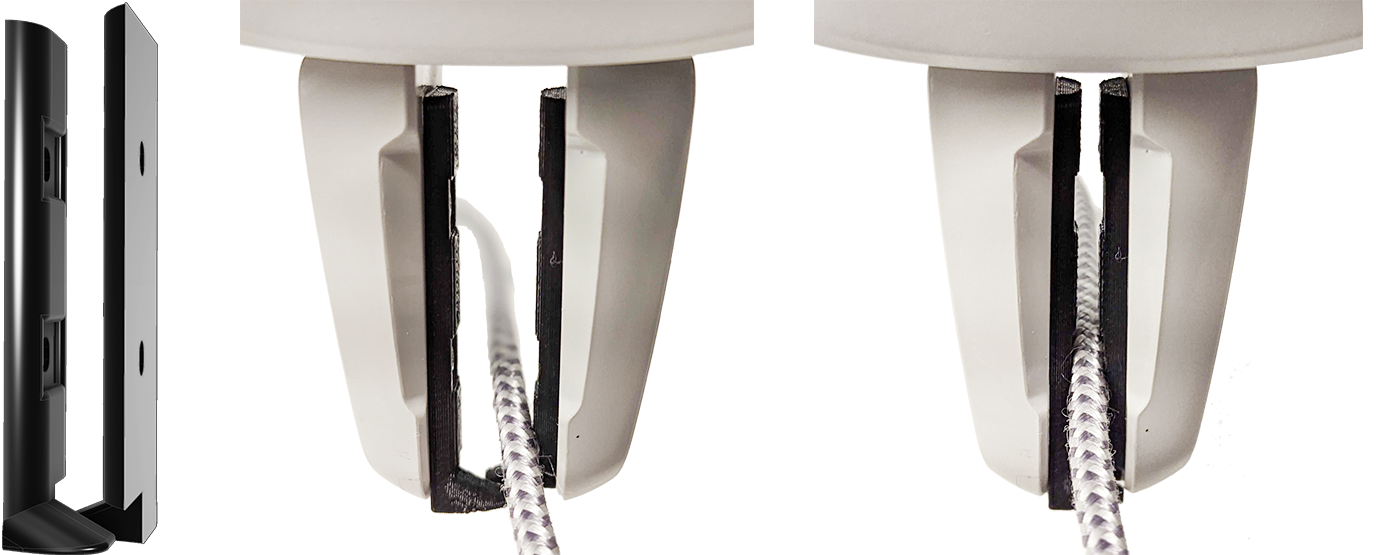}
\caption{\textbf{Pinch-Cage (PC) Jaws}. Left: a rendering of the design of the pinch-cage gripper jaws, which attach to the YuMi's standard white gripper. Center: the cage grasp, in which the perpendicular "foot" segment of the attachment enables the cable to slide freely. Right: the pinch grasp, in which the cable is held tightly, preventing slack from slipping through.
}
\label{fig:grippers}
\end{figure}

\label{sec:methods}
\algabbr{} (Figure~\ref{fig:algorithm}) proceeds in iterations consisting of two phases. In the first phase, \algabbr{} actively manipulates the cable to identify knots by pulling apart the endpoints, shaking the cable, and tracing along the cable physically. In the second phase, it identifies exposed knots and undoes them using a novel bilateral manipulation primitive that cages the knot and slides it apart. In this section we will describe \textit{1)} the novel gripper jaw design, which allows multiple grasping modes with a compact form factor and no additional moving parts; \textit{2)} the manipulation primitives used as sub-components in \algabbr{}; \textit{3)} the perception systems used, and \textit{4)} how \algabbr{} uses these components to untangle long cables.

\subsection{Pinch-Cage (PC) Grippers}

Long cables can limit the efficacy of pinch grasps, in which both grippers firmly grasp the cable, because such grasps are unable to efficiently manage slack between grippers. Prior work demonstrates the promise of sliding grippers along cables~\cite{she2019cable}, and to this end we present a passive mechanical design that facilitates pinch-pinch, cage-cage, and cage-pinch grasps. The design, shown in Fig.~\ref{fig:grippers}, attaches to the YuMi robot's standard parallel jaw grippers. Each jaw includes a perpendicular ``foot" segment that facilitates both \textbf{caging grasps}, where the jaws open partially and the feet prevent the cable from slipping out, and \textbf{pinching grasps}, where the jaws firmly grasp the cable. Caging grasps enable the cable to freely slide inside the gripper, while pinch grasps impart more force and prevent slack from slipping through. These two modes of cable grasping enable a variety of manipulation primitives that are particularly suitable for long cables, without the need for sensing or closed-loop control. 


\subsection{Manipulation Primitives}
\label{subsec:primitives}
\algabbr{} \looseness=-1 employs 5 cable manipulation primitives which are facilitated by the PC jaws and use an analytic grasp planner which selects collision-free gripper poses such that the grippers avoid each other as well as neighboring cable segments.

\subsubsection{Reidemeister move}
To uncover the cable's underlying knot structure $\mathcal{K}(c_t)$, the robot spreads it apart by pinching both of the endpoints, $c_t(0)$ and $c_t(1)$, and pulling them 1.2 meters apart towards opposite ends of the workspace. This was determined empirically to remove occlusions while ensuring no knots are tightened. This is an example of a Reidemeister move in knot theory, and in this case it serves to spread out slack to reveal knots.

\subsubsection{Cable Shaking}
The Reidemeister move requires both cable endpoints to be visible and graspable. When this condition does not hold, we leverage dynamic shaking actions, a popular manipulation primitive in deformable manipulation~\cite{ha2022flingbot, zhang2018deep}. In this paper, the robot performs shaking actions to uncover occluded endpoints and loosen knot structure. If no endpoints are identifiable, the robot computes a random point on the cable's mask in the overhead RGBD image. Using the pointcloud from the RGBD image, the robot identifies and pinches the 3D point corresponding to this point, $c_{\rm shake}$. After pinching, the robot raises its arm 0.7 meters off the table and executes a shaking motion by rotating the wrist joint 3 times by 2 radians, at a frequency of 1.5Hz, with a radius of 0.1 meters. These parameters were determined empirically to maximize the disruption to the cable's visual state in consideration of the YuMi robot's limits.
If exactly one endpoint is visible and graspable, and the previous move was not a shaking action, the robot pinches this endpoint, $c_{\rm shake} = c_t(0)$ or $c_{\rm shake} = c_t(1)$.

We also make use of the shaking action as a recovery move in case of any failures that may occur in other primitives.

\subsubsection{Bimanual physical tracing}
During bimanual physical tracing, the robot slides along the cable until it visually identifies either an endpoint or a knot. 
In this motion, the robot first pinches two nearby points on the cable $c_t(s_{\rm trace, l})$ and $c_t(s_{\rm trace, r})$ with the left and right arms respectively. Without loss of generality, let the left arm be the \textit{pulling arm} and the right arm be the \textit{sliding arm.} The robot alternates between pinching and caging between the two arms to slide the cable through one gripper with the other. The robot first cages the cable with the sliding arm while the pulling arm pinches and pulls the cable through by 0.1 meter segments as shown in Fig \ref{fig:sliding_stop_cond}. After each segment, the sliding arm pinches the cable and the pulling arm cages the cable, allowing it to move forward to meet the sliding arm without dropping or pushing the cable. The robot repeats this procedure until it either detects an endpoint or a knot approaching the sliding arm from overhead images. The detection algorithms are described in Section~\ref{subsec:perception}.

While sliding along the cable, the robot dangles the cable 0.45 meters above the work surface, causing loose loops to fall away while knots remain. \algabbr{} uses this behavior to closely inspect the cable to actively locate knots in the presence of excess slack.

This motion begins at an endpoint, so that during execution the portion of cable that has been physically traced contains no knots, an invariant useful for managing slack in subsequent steps and indicating untangling success. If the robot slides uninterrupted from one endpoint to the other without detecting knots, it can verify that the entire cable contains no more knots. This is used as a termination condition by \algabbr{}.


\subsubsection{Knot isolation}
Immediately after physical cable tracing terminates at a knot, we execute this primitive, which improves the likelihood of successful untangling in future steps. The sliding arm deposits the knot it is holding onto the workspace, while the other arm sweeps aside the remaining slack not pertaining to the knot.
\subsubsection{Dual-cage separation}
To untie individual knots, the robot attempts to cage two points inside the knot and then slowly pull its arms apart while wiggling the wrist joint 15 times by 0.2 radians. This wiggling helps reduce friction between segments of cable sliding past each other. Precisely, given a knot $[a, b]$ in cable $c_t$, the robot grasps two graspable points $c_t(s_{\rm knot, l})$ and $c_t(s_{\rm knot, r})$ with the left and right arms respectively, such that $s_{\rm knot, l}, s_{\rm knot, r} \in (a, b)$.
The double cage grasps allow the cable to freely slip through the fingers as the knot loosens, allowing the robot to untie knots in one action where the arms move as far apart as kinematically feasible, approximately 1 m apart. If the robot encounters sufficient resistance due to an endpoint or knot at either gripper, exceeding the YuMi torque limits, our algorithm will automatically stop and reset.



\subsection{Perception Systems}\label{subsec:perception}
In contrast to some prior work which performs full state estimation of the cable~\cite{lui2013tangled}, we rely on learned perception methods, since the significant lengths of slack introduce occlusions that make full state estimation difficult. In addition, to initiate certain manipulation primitives in Section~\ref{subsec:primitives}, \algabbr{} also relies on additional manipulation primitives to expose key areas of the cable, such as knots, to predict task-relevant untangling keypoints.

\begin{figure}[t!]
\centering
\includegraphics[width=1.0\linewidth]{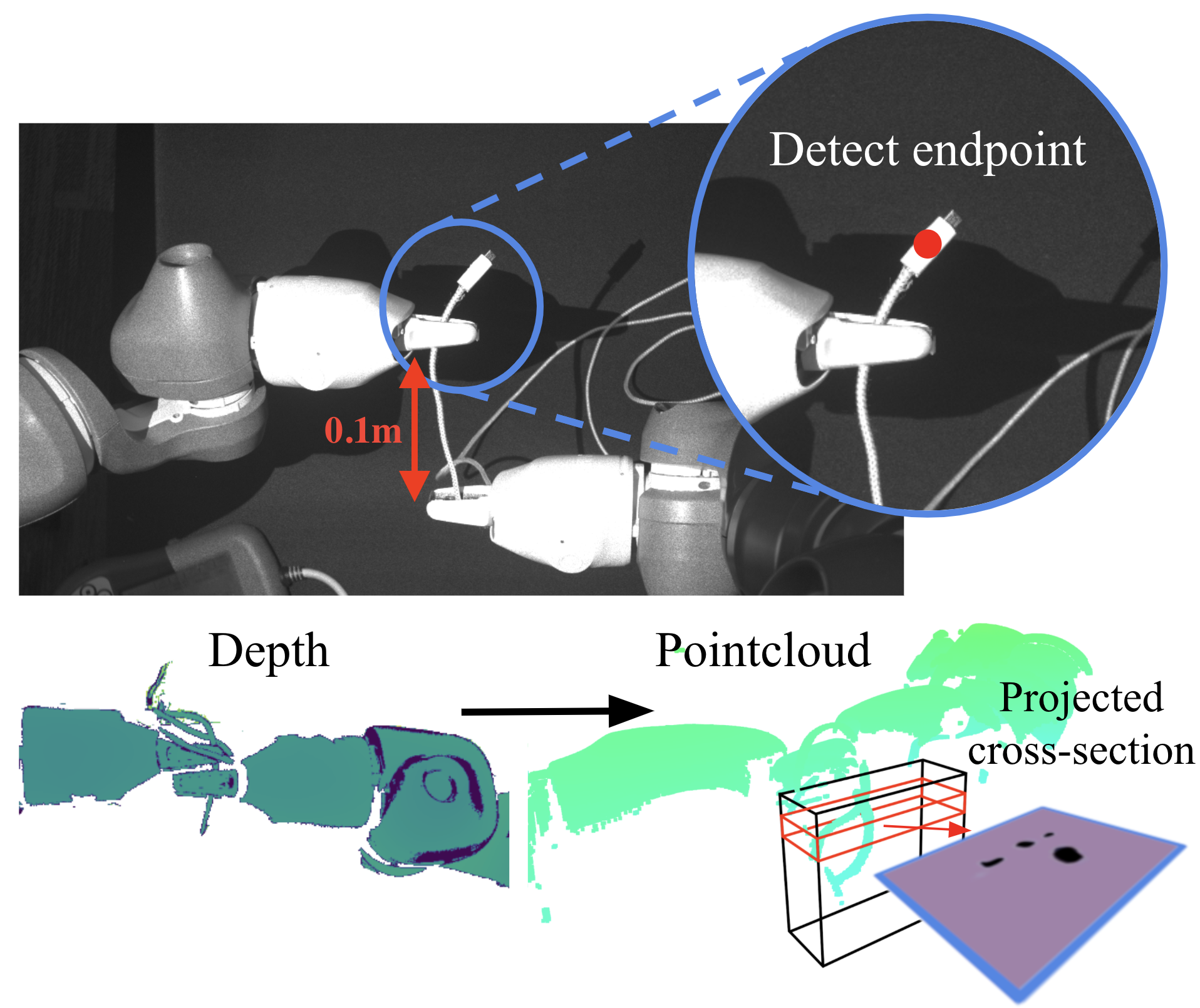}
\caption{\textbf{Physical Tracing Stop Condition}: \textit{Top}: Physical cable tracing uses an endpoint detector to locate endpoints and terminate when any reach close to the gripper. \textit{Bottom}: Physical cable tracing detects knots by horizontally slicing the pointcloud in front of the gripper (right), then analyzing the cross-sections for the number of connected components. In this case, the figure-eight knot contributes 4 connected components (black) to a cross section and reaches the threshold for number of points, so a knot is detected. If neither an endpoint nor a knot is detected, the robot continues sliding by beginning at the position shown in the bottom image, caging the left gripper and pinching the right gripper, and pulling the right gripper back as in the top image, tracing along the cable until a knot or endpoint is reached. 
}
\label{fig:sliding_stop_cond}
\end{figure}

\subsubsection{Endpoint detection}\label{subsubsec:endpoints}
In the Reidemeister move and bimanual physical tracing primitives, \algabbr{} relies on detecting endpoints in the image. We collect and manually label a dataset of 700 images in the workspace of the cable either on the manipulation surface or in the gripper jaws during physical tracing. We train a fully-convolutional network~\cite{long2015fully} based on a ResNet 50 backbone to output a Gaussian heatmap over each endpoint as in~\cite{grannen2020untangling,sundaresan2021untangling,viswanath2021disentangling}.

\subsubsection{Visual cable tracing}\label{subsubsec:alg_cable_tracing}
\begin{figure}[t!]
\centering
\includegraphics[width=0.7\linewidth]{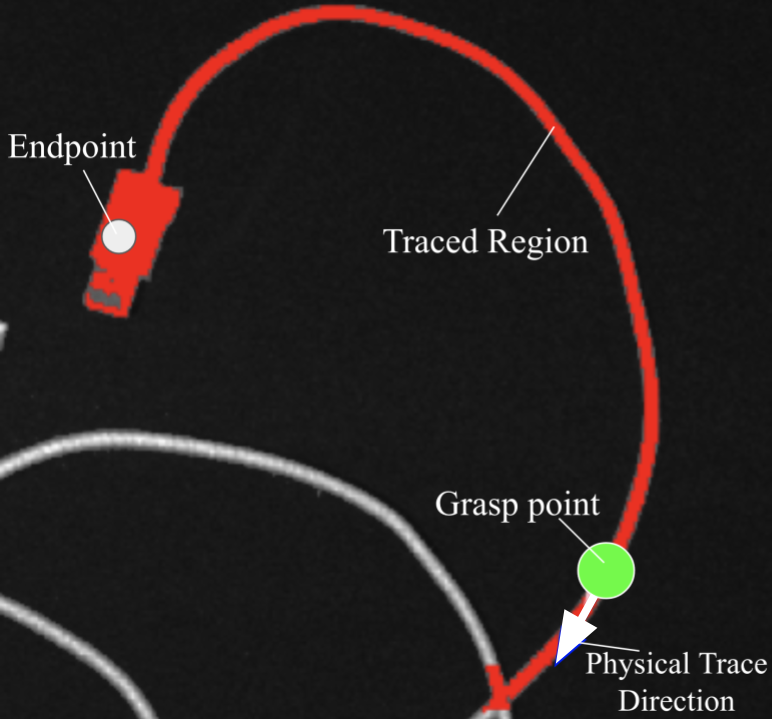}
\caption{\textbf{Visual Cable Tracing}: Beginning from the detected endpoint keypoint (white), the algorithm traces the cable using BFS along cable pixels until it detects that it has reached a crossing, after which it backtracks to return the grasp point (green) and slide direction (blue) for subsequent physical cable tracing.
}
\label{fig:tracing}
\end{figure}

SGTM uses \emph{visual} cable tracing to identify good starting points for \emph{physical} tracing (Figure~\ref{fig:tracing}). Given an RGB image of a segmented cable and an endpoint pixel location, algorithmic cable tracing uses a modified breadth-first search (BFS) to follow the cable mask in the image and stop at the first self-crossing. It detects crossings by monitoring the frontier of added pixels at each BFS iteration. If the bounding box tightly surrounding the frontier has side length over $12 \mathrm{px}$, the algorithm terminates, as the search is now bleeding across different cable segments. After termination, the algorithm backtracks by $100 \mathrm{px}$ and returns a pixel location along the cable which is a safe distance from the crossing to grasp. \algabbr{} uses this keypoint to expedite bimanual physical cable tracing by avoiding physically tracing segments from the endpoint prior to any crossings.

\subsubsection{Physical tracing stopping condition}
During physical cable tracing, \algabbr{} pulls the cable through the sliding gripper in segments of 0.1 meters. After every segment, the stopping condition categorizes the next segment of cable near the sliding gripper as a knot, endpoint, or straight cable in the following manner: the robot takes an overhead image and evaluates the endpoint detector (Section~\ref{subsubsec:endpoints}) to check if any endpoints are within a $0.1\times0.1\times0.1m$ cube around the sliding arm grippers by using the depth values at each predicted endpoint. In order to suppress false positives, we ignore endpoint detections when over $0.4 m$ of the cable remains to be traced. If an endpoint is found after this range, the stopping condition returns \textit{ENDPOINT}. This is shown in Fig \ref{fig:sliding_stop_cond}.s

\begin{figure*}[!htbp]
\centering
\includegraphics[width=1.0\linewidth]{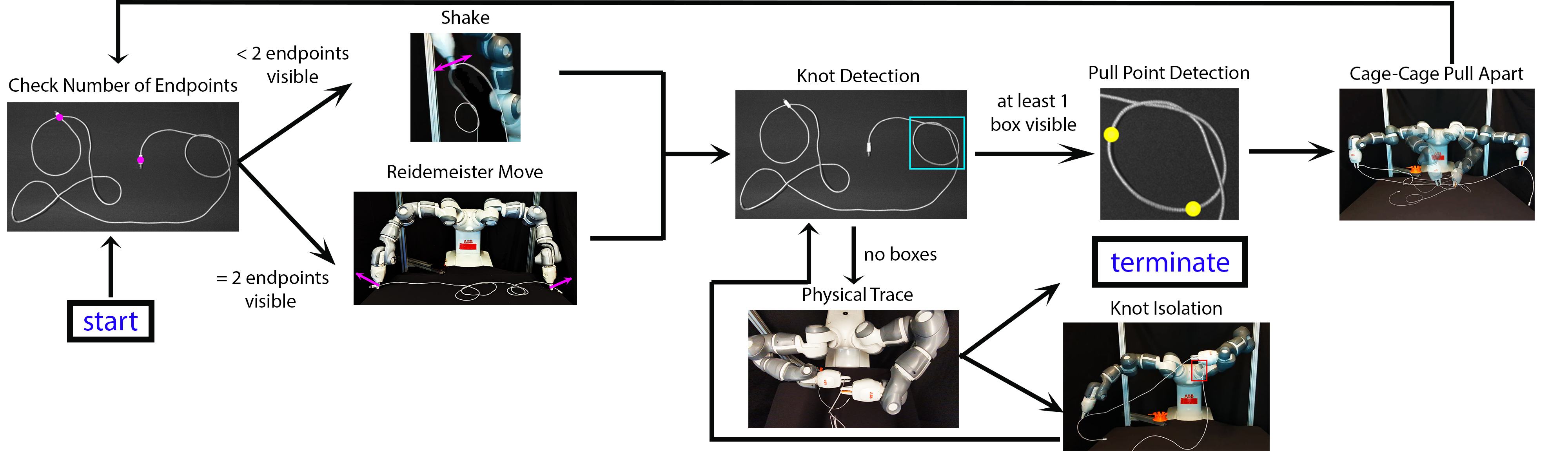}
\caption{\textbf{\algname{}}: \algabbr{} begins by recognizing endpoints. If both are visible, it proceeds with a Reidemeister move. If not, it shakes. Afterwards, knot detection is performed after which it untangles if a knot is visible. If not, it performs physical tracing to ensure no knots remain. If a knot is found while tracing, the knot is isolated and placed in the workspace, and the algorithm returns to the knot detection step.
}
\label{fig:algorithm}
\end{figure*}
If an endpoint is not found, the robot analyzes the pointcloud surrounding the gripper of the sliding arm to categorize whether the next segment of cable is a knot or a straight cable. The robot performs a $3\times13\times6$ cm volume crop in front of and beneath the sliding gripper to capture the next segment of cable. This volume is separated into 1 cm thick horizontally-sliced cross sections, which are each analyzed for the number of connected components in a re-projected depth image (see Fig.~\ref{fig:sliding_stop_cond}). If any cross section contains multiple connected components, this could either indicate a loop that has not been undone by gravity during physical tracing or a knot. If any cross section contains multiple connected components and the number of points in the volume crop is at least 1000 points, the stopping condition returns \textit{KNOT.} If no cross sections contain multiple connected components, a knot could still exist if it is very tightly zipped together or self-occluded. Thus, if the volume crop contains at least 2000 points, the stopping condition also returns \textit{KNOT.} If none of these conditions are satisfied, the stopping condition returns \textit{STRAIGHT}.

\subsubsection{Knot detection}
\looseness=-1 We train an object detection network, based on the Mask R-CNN architecture and implemented using the Detectron2 codebase~\cite{wu2019detectron2}, to detect and classify figure-eight and overhand knots in cable images taken by the overhead PhoXi camera.
The network is initialized with weights from a ResNet-50 FPN backbone pretrained on the COCO dataset~\cite{lin2014microsoft} and trained on a dataset of 312 images with manually annotated knots in images containing both knots and loops.

\subsubsection{Pull point detection}
\label{HULK}
\begin{figure}[t!]
\centering
\includegraphics[width=1.0\linewidth]{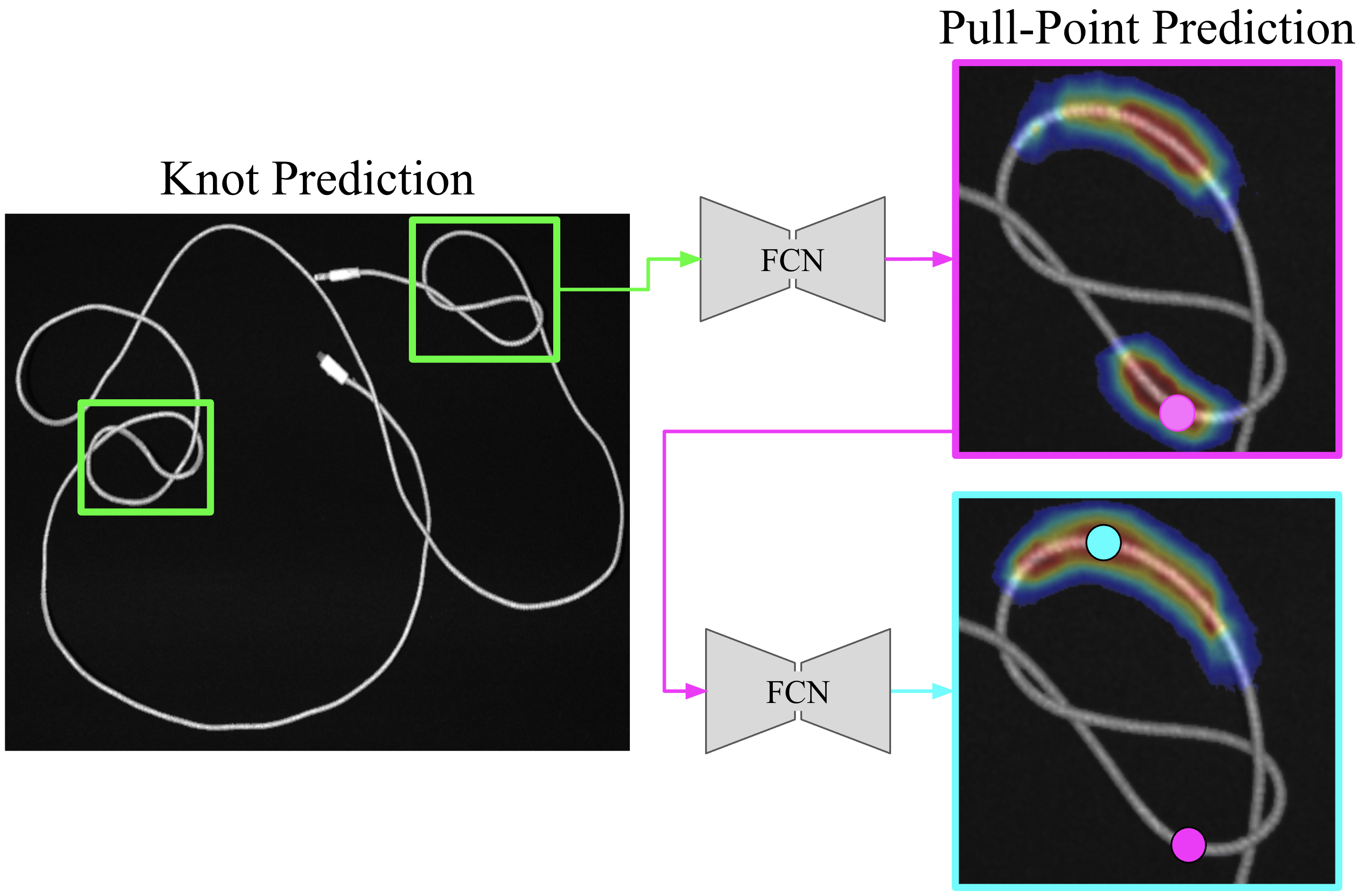}
\caption{\textbf{Learned Pull Point Prediction}: Left: predictions from the knot detection method on an image containing two figure-eight knots. Right: visualization of the cascading pull point detector. The top network outputs a heatmap over potential grasp points, and we select the point at which the heatmap has the highest value. The bottom network is conditioned on the first selected pull point (magenta) to predict a heatmap for the second pull point (blue). 
}
\label{fig:cascade}
\end{figure}

If the knot detection step locates a knot bounding box, it outputs a crop of the knot, which is passed through a two-stage cascading network. 
Both independently-trained stages use a Resnet34 backbone followed by a sigmoid activation to predict two pull points for performing a dual-cage separation action. The first network outputs a heatmap for both potential pull points. Of these, we select the location of the highest heatmap value, projected onto the nearest cable segment. The second network, trained separately, takes in the same cropped image with a fourth channel encoding a heatmap centered around the already-chosen first point, and it outputs a heatmap from which the second point is chosen the same way, using an argmax. This process is visualized in Figure~\ref{fig:cascade}.

We use the two-network cascading design because with a single network, we find that pick point multimodality often causes heatmap outputs to bleed together significantly, making it difficult to choose two distinct points. 
Conditional action prediction has also been previously studied in deformable manipulation by~\citet{wu2019learning} for cable and fabric manipulation.

\subsection{\algname{}}
\label{sec:alg}

We now describe the \algname{} algorithm, which combines the manipulation primitives and perception subsystems from Sections~\ref{subsec:primitives} and~\ref{subsec:perception} to untangle long cables while managing cable slack. An overview of the algorithm is displayed in Fig. \ref{fig:algorithm}. During each iteration, \algabbr{} alternates between two phases: physical manipulation to increase knot visibility (\textit{Active Knot Perception}) and predicting keypoints for a) the dual-cage separation maneuver to loosen and untie knots once they are detected or b)  physically checking if the cable is untangled when no knots are detected (\textit{Knot Untangling/Physical Tracing}).

\begin{enumerate}
    \item \textit{Active Knot Perception:} During execution, at each iteration, \algabbr{} begins by detecting visible endpoints. If both are detected, it executes a Reidemeister move, and if none are detected it executes a shake action from the centroid of the cable cluster. If a shake action was previously performed and no endpoints are visible still, the shake action is executed again from a random location on the cable. All three types of moves serve to spread out the cable and increase the chance of perceiving knots. 
    \item \textit{Knot Untangling/Physical Tracing:} Next, if knot bounding boxes are detected, the robot executes a dual-cage separation action to attempt to untie the knot closest to an endpoint (found using the technique in \ref{subsubsec:alg_cable_tracing}) based on grasp keypoints the perception system outputs. If \algabbr{} does not detect knots, the robot initiates the physical tracing stage to confirm that the cable is untangled, either ending in termination or by detecting a knot and isolating it on the table for further untangling.
\end{enumerate}


%% file: 5-experiments.tex
\section{Physical Experiments}
\label{sec: experiments}

We evaluate \algabbr{} on a set of physical cable untangling experiments using the bilateral ABB YuMi robot. The experiments evaluate whether \algabbr{} can untangle different classes of initial configurations and terminate when it has completely untangled the cable.

\subsection{Difficulty Tiers and Performance Metrics}

\begin{figure}[t!]
\centering
\includegraphics[width=1.0\linewidth]{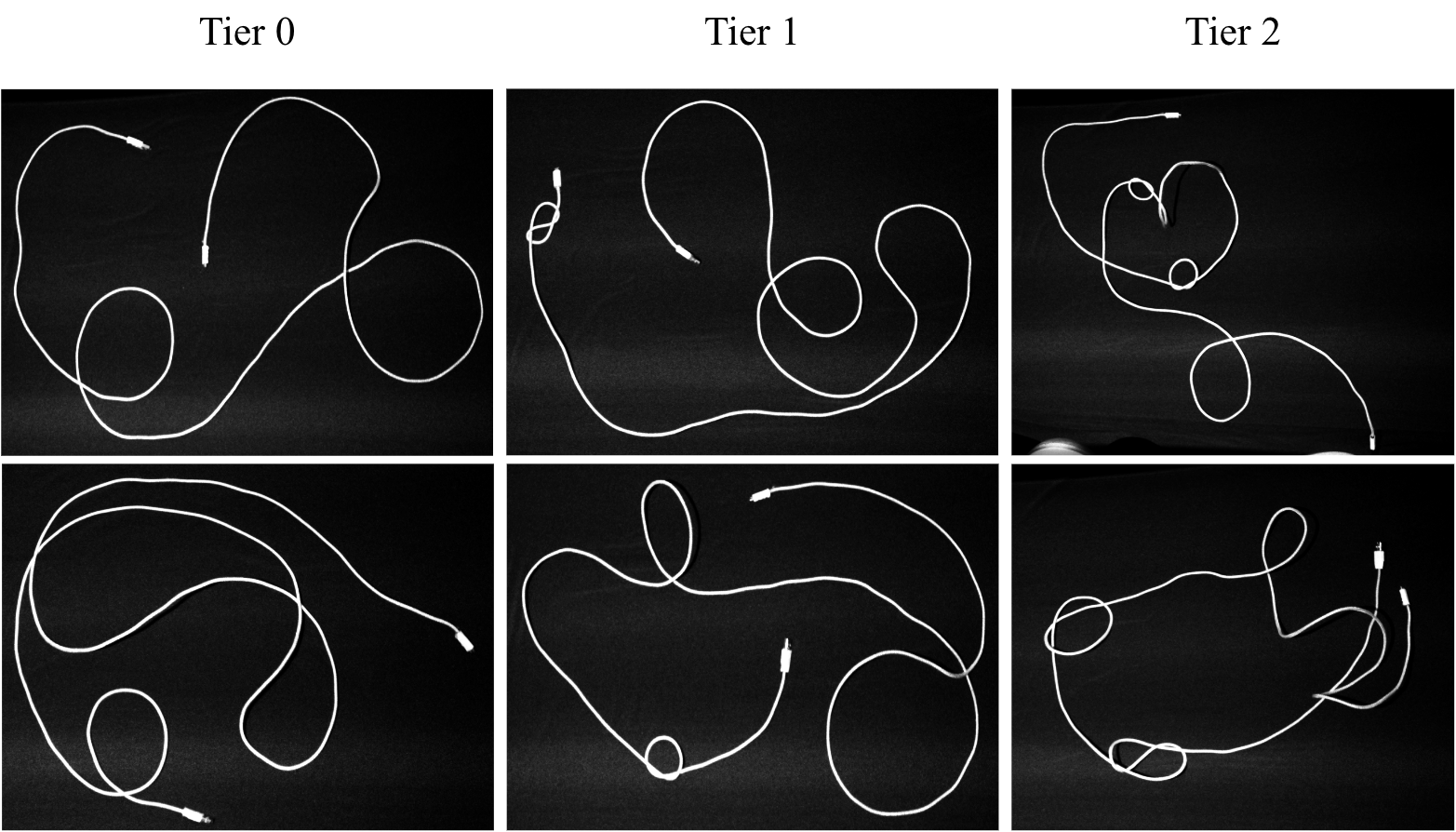}
\caption{\textbf{Cable Configurations}: Shown here are example configurations of the cable across the 3 tiers we run experiments on. 
}
\label{fig:start_configs}
\end{figure}

We consider several methods for initializing the cable state, with examples shown in Fig. \ref{fig:start_configs}. These are categorized into the following tiers of difficulty:
\begin{itemize}
    \item \textbf{Tier 0}: Cable has no knots, but the slack is piled randomly within the workspace by holding both endpoints 1.5 meters above the workspace and dropping them. In this tier, the robot must successfully verify that the cable contains no knots before terminating. We report the number of trials that successfully detect that the cable is untangled (Untangled Detection Rate), average number of manipulation actions (Avg. Number of Actions), and average time to detect that the cable is untangled (Avg. Untangled Detection Time).
    \item \textbf{Tier 1}: Cable has a single overhand or figure-eight knot that is loose (12-14 cm diameter) or tight (6-8 cm diameter), located close to an endpoint (side), 0.75 meters from the endpoint (mid-center), or 1.5 meters from an endpoint (center). It is arranged with the knot isolated from the rest of the cable slack, and the slack is randomized by lifting both endpoints as high above the workspace as possible without lifting the knot itself from the workspace and dropped. We report the number of trials that successfully untangled the knot (Untangling Success Rate), the number of trials that both untangled the knot and detected that the knot was untangled (Untangled Detection Rate), the average time taken to untangle the knot (Avg. Time to Untangle), and the average time taken to successfully identify that the cable was untangled (Avg. Untangled Detection Time).
    \item \textbf{Tier 2}: Cable has two overhand or figure-eight knots (including mixed types) that are loose or tight, in series, located close to each other ($<$ 0.75 m apart) or far from each other ($>$ 1.5 m apart). Similar to tier 1, both knots are isolated from the rest of the cable slack and the endpoints are lifted as high above the workspace as possible without lifting either knot from the workspace and dropped. We record the number of trials that successfully untangle a single knot (Untangling 1 Success Rate), number of trials that successfully untangle both knots (Untangling 2 Success Rate), the number of trials that untangle both knots and detect that it is untangled (Untangled Detection Rate), average number of manipulation actions (Avg. Number of Actions), average time to untangle the first knot (Avg. Time to Untangle 1), average time to fully untangle the cable (Avg. Time to Untangle 2), and the average time to untangle both knots and detect that the cable is untangled (Avg. Untangled Detection Time).
\end{itemize}

For each tier 1 or 2 experiment, we record the average time until the cable has been successfully untangled. Because the robot may successfully untangle the cable but not successfully recognize that the cable is untangled (Section~\ref{subsec:failure_modes}), we separately record the average time to successfully detect the cable is untangled (when this is the case). We cap each experiment to 15 minutes in tiers 0 and 1 and to 20 minutes in tier 2. This is because tier 2 configurations are much more difficult and require more time to untangle. We analyze the observed failure modes of the algorithm in Section~\ref{subsec:failure_modes}.

\subsection{Experimental Setup}
\label{subsec:experimental_setup}
The workspace contains a bimanual ABB YuMi robot with two PC grippers. The manipulation surface is black and foam-padded to avoid end effector damage during any workspace collisions. The cable is a light-gray, braided 2.7 m micro-USB to USB cable that can be segmented from the manipulation surface via color thresholding. The workspace has an overhead PhotoNeo Phoxi Camera that captures depth and grayscale images of resolution 732 x 1142\textit{px}. 

\begin{table}[!t]
    \centering
    \begin{tabular}[width=1.0]{||c|c|}
    \hline
    Untangled Detection Rate &  8/12\\
    \hline
    Avg. Number of Actions &  4.33\\
    \hline
    Avg. Untangled Detection Time (s) &  266\\
    \hline
    Failures &  A (1), B (1), C (2), D (0), E (0)\\
    \hline
    \end{tabular}
    \caption{\normalfont{\textbf{Tier 0 results: }In 12 tier 0 experiments, \algabbr{} must detect that a cable with no knots is untangled. We observe that \algabbr{} is able to successfully terminate in 8/12 cases using an average of 4.33 manipulation primitive actions. See Section~\ref{subsec:failure_modes} for analysis of the failure modes A-E.}}
    \label{tab:results_tier0}
\end{table}


\begin{table}[!t]
    \centering
    \begin{tabular}[width=1.0]{||c|c|c|}
    \hline
    & \textbf{Loose} & \textbf{Dense}\\
    \hline
    Untangling Success Rate &  4/6 & 4/6\\
    \hline
    Untangled Detection Rate &  1/4 & 2/4\\
    \hline
    Avg. Number of Actions & 7.17 & 7.5\\
    \hline
    Avg. Time to Untangle (s) &  154 & 139\\
    \hline
    Avg. Untangled Detection Time (s) &  270 & 527\\
    \hline
    Failures &  A (1), B (0),  & A (1), B (0), \\
    & C (2), D (0), & C (1), D (1),\\
    & E (0), F (2) & E (0), F (1)\\

    \hline
    \end{tabular}
    \caption{\normalfont{\textbf{Tier 1 results:} In 12 experiments, the cable starts off with a single knot in its initial configuration. \algabbr{} is able to successfully untangle the cable 4/6 times in both the loose and dense cases of this problem. However, when the robot untangles the cable, it often fails to detect that it has successfully untangled the cable. This occurs because the robot forms loops that are not knots during execution, which are mistaken for knots during physical tracing (Failure Mode C).}}
    \label{tab:results_tier1}
\end{table}


\begin{table}[!t]
    \centering
    \begin{tabular}[width=1.0]{||c|c|c|}
    \hline
    & \textbf{Loose} & \textbf{Dense}\\
    \hline
    Untangling 1 Success Rate &  5/6 & 5/6\\
    \hline
    Untangling 2 Success Rate &  3/6 & 3/6\\
    \hline
    Untangled Detection Rate & 0/3 & 1/3\\
    \hline
    Avg. Number of Actions &  9.83 & 10.17\\
    \hline
    Avg. Time to Untangle 1 (s) & 71 & 112\\
    \hline
    Avg. Time to Untangle 2 (s) & 475 & 670\\
    \hline
    Avg. Untangled Detection Time (s) & N/A & 1079\\
    \hline
    Failures &  A (0), B (1),  & A (1), B (1), \\
    & C (1), D (0), & C (1), D (1),\\
    & E (1), F (3) & E (1), F (0)\\
    \hline
    \end{tabular}
    \caption{\normalfont{\textbf{Tier 2 results:} In this tier of 12 experiments for dense and loose knots, the cable starts off with two knots in its initial configuration. The robot is able to successfully untangle at least one of the knots 5/6 times in both the loose and dense versions of this tier. In 3/6 of both the loose and dense cases, \algabbr{} successfully untangles both knots. However, similar to Tier 1, many loops and sometimes new knots are formed during untangling, which lead to timing out during a trial due to failure of untangled detection.}}
    \label{tab:results_tier2}
\end{table}


\subsection{Results}
\looseness=-1 In tier 0, \algabbr{} successfully detects that 8/12 cases are untangled, taking an average of 4.33 manipulation primitive actions and 266 seconds (Table~\ref{tab:results_tier0}). In the cases where \algabbr{} does not terminate successfully, the robot either moves the cable off the manipulation surface, creates a new knot during execution, or falsely detects loops as knots during physical tracing.

In tier 1, \algabbr{} successfully untangles the cable in 4/6 trials in both the loose and the dense categories (Table~\ref{tab:results_tier1}). The untangled detection rate is 1/4 for loose configurations and 2/4 for dense configurations, and the most common errors are detection of loops as knots during physical tracing and system errors with the YuMi robot.

In tier 2, \algabbr{} is able to untangle a single knot in 5/6 trials in both the loose and dense cases (Table~\ref{tab:results_tier2}). In both cases, the robot untangles both knots 3/6 times. The robot often has trouble detecting that it has successfully untangled the cable in this case, due to an accumulation of twists during untangling that lead to loops that look like knots during physical tracing.


\subsection{Failure Modes}\label{subsec:failure_modes}

\begin{figure}[t!]
\centering
\includegraphics[width=1.0\linewidth]{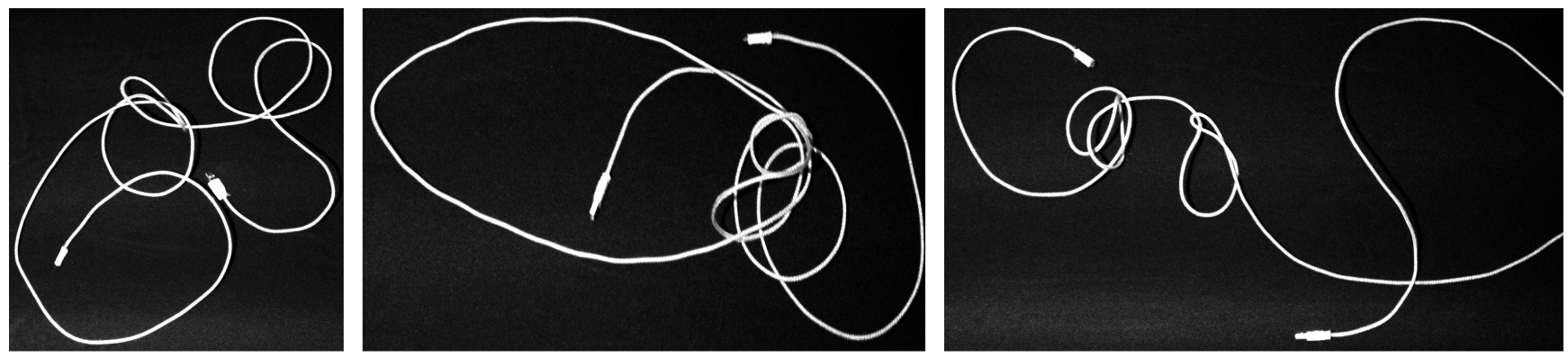}
\caption{\textbf{Failure Mode A Examples}: These are examples of complex knots or unseen knots that form while untangling, but fall out of the distribution of training samples for the knot detector, causing a failed knot detection.
}
\label{fig:failures_fig}
\end{figure}

In experiments for \algabbr{}, we observe the following failure modes:
\begin{enumerate}[(A)]
    \item Unseen or complex knots form while untangling and are not detected by Mask R-CNN for knot bounding box detection.
    \item The cable falls out of the reachable workspace.
    \item Repeated false positive knot detections during physical tracing result in hitting the time limit.
    \item Missed knots during physical cable tracing cause knots to tighten.
    \item Knot untangled, but not enough time to check termination.
    \item YuMi robot system or reset errors.
\end{enumerate}
(A) During execution, \algabbr{} sometimes introduces new knots by pulling endpoints through existing loops on top of them as shown in Figure~\ref{fig:failures_fig}. Though unlikely, these occurrences are often difficult to recover from since knots formed in this way are visually distinct from the training set of knot bounding box detection model. This failure happens particularly often after no endpoints are detected and the robot must shake from a random cable location, after which endpoints are buried underneath slack.

(B) Because we shake near the edge of the table, sometimes the motion causes part of the cable to drop off the side, which can cause the rest of the cable to follow after it is released.

(C) One particularly confusing case for knot detection during physical tracing is tightly twisted loops which do not fall away due to gravity. Sometimes this happens multiple times in a rollout, wasting time and eventually causing the time limit $T_{\rm timeout}$ to be reached.

(D) If \algabbr{} fails to detect a knot during physical cable tracing, the sliding actions can tighten the knot until it is too dense to fit grippers inside, resulting in failure.

(E) Sometimes, \algabbr{} is able to recover from situations like failure mode (A), but this can happen very late in the experiment, resulting in not enough time to check termination. 

(F) YuMi errors result from 1) the robot getting stuck at a singularity from which it is unable to reset or 2) the robot attempting to reach a point just outside of its reachable workspace and producing errors from which it cannot recover.

\looseness=-1We observe that the physical tracing termination condition is low-recall but high-precision; that is, if it claims that the cable is untangled, it most likely is (only in one case out of 36 experiments did it output a false positive). However, while false positives terminate by presenting a tangled cable as untangled, a false negative results in the robot continuing to try to untangle an untangled cable. The latter is less severe, because the robot often eventually discovers that the cable is untangled; thus, we prefer \algabbr{} to err on the side of false negatives.


%% file: 6-conclusion.tex
\section{Discussion and Future Work}
For untangling a single long (up to 3m) cable, this work presents a formal problem definition, a novel jaw design, novel perception-based primitives, and \algname{}, an algorithm for untangling long cables. \algabbr{} introduces 3 novel manipulation primitives for cable manipulation: shaking, physical tracing, and dual-cage separation. These primitives aid in easing perception and managing slack in long cables. \algabbr{} also introduces perception systems that guide the usage of these primitives. Experiments show that \algabbr{} can untangle long cables with a 58.3\% success rate overall across tiers containing one to two knots. 


\looseness=-1 Future work will investigate refining these primitives to improve performance, generalizing this method to less structured starting configurations of cables and knots, as well as multiple cables. Further, active perception for cable manipulation is of interest, for example investigating ways of interrogating the presence of knots with finer precision via high frame rate videos and gripper perturbations. In addition, replacing the manually tuned components of \algabbr{}, for example the knot detection threshold, with more data-driven methods and self supervised data collection may increase robustness and generalization.

%% file: 7-acknowledgements.tex
\section*{Acknowledgments}
\footnotesize
This research was performed at the AUTOLAB at UC Berkeley
in affiliation with the Berkeley AI Research (BAIR) Lab, the
CITRIS “People and Robots” (CPAR) Initiative, and the RealTime Intelligent Secure Execution (RISE) Lab. The authors were
supported in part by donations from Toyota Research Institute
and by equipment grants from PhotoNeo, and Nvidia.